\documentclass[journal, 10pt]{IEEEtran}

\ifCLASSINFOpdf
\else
\fi
\usepackage{cite}
\usepackage{url}
\usepackage{hyperref}
\usepackage{graphicx}
\usepackage{float}
\usepackage[small]{caption}
\usepackage{subcaption}
\usepackage{color,soul}
\usepackage{array}
\usepackage{amsmath}
\usepackage{booktabs}
\usepackage{footnote}
\makesavenoteenv{tabular}
\makesavenoteenv{table}
\usepackage{color}
\usepackage{amssymb}
\usepackage[linesnumbered,ruled,vlined]{algorithm2e}
\usepackage[noend]{algpseudocode}
\usepackage{setspace}
\usepackage{mathrsfs}
\usepackage{mathtools,xparse}
\newtheorem{theorem}{Theorem}[section]

\usepackage{setspace}

\newcommand\tab[1][1.5cm]{\hspace*{#1}}
\begin{document}
\bstctlcite{IEEEexample:BSTcontrol}
%\pagenumbering{gobble}
%\title{AISYN: AI-driven Logic Synthesis Framework}
\title{AISYN: AI-driven Reinforcement Learning-Based Logic Synthesis Framework}

\author{\IEEEauthorblockN{Ghasem Pasandi, Sreedhar Pratty, and James Forsyth}\\
\IEEEauthorblockA{\textit{NVIDIA Corp.} \\
Santa Clara, CA\\
\{gpasandi, spratty, jforsyth\}@nvidia.com}

}

% after the title page when using the two side option.
%

\maketitle

\begin{abstract} 
Logic synthesis is one of the most important steps in design and implementation of digital chips with a big impact on final Quality of Results (QoR). For a most general input circuit modeled by a Directed Acyclic Graph (DAG), many logic synthesis problems such as delay or area minimization are NP-Complete, hence, no optimal solution is available. This is why many classical logic optimization functions tend to follow greedy approaches that are easily trapped in local minima that does not allow improving QoR as much as needed. We believe that Artificial Intelligence (AI) and more specifically Reinforcement Learning (RL) algorithms can help in solving this problem. This is because AI and RL can help minimizing QoR further by exiting from local minima. Our experiments on both open source and industrial benchmark circuits show that significant improvements on important metrics such as area, delay, and power can be achieved by making logic synthesis optimization functions AI-driven. For example, our RL-based rewriting algorithm could improve total cell area post-synthesis by up to 69.3\% when compared to a classical rewriting algorithm with no AI awareness. 
\end{abstract}
%%%%%%%%%%%%%%%%%%%%%%%%%%%%%%%%%%%%%%%%%%%%%%%%%%%%%%%%%%%%%%%%%%%%%%%%%%%%%%%%%%%%%

\IEEEpeerreviewmaketitle

\section{Introduction}
\label{Intro:sec}
\setstretch{0.98}
Artificial Intelligence (AI) is the single most powerful driving force of our time. AI has already started changing our daily lives, fueled many research projects, helped improve the quality of many engineering products, and enabled people to rethink how we integrate information, analyze data, and use the resulting insights to improve the decision-making \cite{AI_transform}. This is why it is easy to see the fingerprint of AI in intelligent applications ranging from image classification and recognition \cite{lu2007survey, krizhevsky2017imagenet}, virtual assistants (e.g. Alexa and Siri), autonomous vehicles \cite{li2020ai, caesar2020nuscenes}, mathematics \cite{li2020fourier}, and more recently in healthcare \cite{vaishya2020artificial, yang2020modified}. Now, most of companies have AI divisions and research teams for AI and there are many startups with a heavy focus on AI or its variations such as Machine Learning (ML), Deep Learning (DL), and Reinforcement Learning (RL). For example, there are over 4,000 AI startups working with NVIDIA corporation, one of the pioneers in AI \cite{AI_single_force}.

There are three pillars for AI to be a powerful computing platform: fast computing engine, data, and algorithm. With the invention of Graphical Processing Units (GPUs) \cite{nickolls2010gpu, keckler2011gpus}, and availability of data in our time, AlexNet \cite{krizhevsky2017imagenet} was the third pillar that ushered AI into a new era. Since then, many deep learning algorithms \cite{srivastava2014dropout}, architectures, and frameworks \cite{he2016deep, szegedy2015going, iandola2016squeezenet} were invented to continuously improve the quality of results by reducing the classification error rate, increasing inference speed \cite{teerapittayanon2016branchynet, han2016eie}, decreasing access to memory \cite{nazemi2019energy}, and so on. This makes AI more advanced than ever before.

With these advancements in AI, now it is time for enabling automation for automation. A good starting point is using AI algorithms to accelerate the chip design workflow  \cite{khailany2020accelerating} or to improve the quality of results generated by the Electronic Design Automation (EDA) tools. There are a few research papers published recently with the focus on using RL/ML/DL algorithms in EDA. For example, Pasandi \textit{et al.} \cite{pasandi2019approximate} formulated the approximate logic synthesis and technology mapping as a reinforcement learning problem and showed that by using simple RL algorithms such as Q-learning \cite{watkins1992q}, significant improvements on Quality of Results (QoR) can be achieved while the error rate is under control. In \cite{yu2018developing, hosny2020drills}, RL/ML algorithms are used to improve the logic and to eliminate the need for a human expert in the logic design process. \textit{Deep-PowerX} \cite{pasandi2020deep} is the most recent attempt to improve power consumption post-synthesis using advances in DL. 

Since many optimization problems in logic synthesis are NP-Complete/NP-Hard, hence, no available optimal solutions, the classical logic synthesis optimization algorithms tend to have a greedy nature which leads to not achieving the best possible results. If we can modify these optimization algorithms in such a way that the chance of getting trapped in local minima is decreased, it will help us to further improve the QoR. We believe that AI and more specifically RL algorithms have this potential. In this paper, we will show that how adding RL smartness to classical logic synthesis optimization functions can result in significant improvements in QoR. More specifically, we will use reinforcement learning and representation learning algorithms to improve a classical logic rewriting optimization function to minimize total cell area post-synthesis. 
%The rest of this paper is organized as follows: Section \ref{level_cyc_sec} presents  finally, Section \ref{conc_sec} concludes the manuscript. 
%%%%%%%%%%%%%%%%%%%%%%%%%%%%%%%%%%%%%%%%%%%%%%%%%%%%%%%%%%%%%%%%%%%%%%%%%%%%%%%%%%%%%
%%%%%%%%%%%%%%%%%%%%%%%%%%%%%%%%%%%%%%%%%%%%%%%%%%%%%%%%%%%%%%%%%%%%%%%%%%%%%%%%%%%%%
\section{Background}
\label{level_cyc_sec}
\subsection{Terminology and Notation}
\label{termin_sub_sec}
\noindent
\textit{G=(V,E):} A Directed Acyclic Graph (DAG) with vertex set of V and edge set of E. \\
\textit{Node or Vertex:} A member of set V.  \\
\textit{n(V)}: Size of set V. \\
\textit{Edge:} A member of set E. \\
\textit{Leaf node:} A node with in-degree of 0. The set of all leaf nodes of G is denoted by $L(G)$. \\
\textit{Root node:} A node with out-degree of 0. The set of all root nodes of G is denoted by $R(G)$.\\
\textit{Internal node:} A node that is not a root or leaf node. \\
\textit{Level of node $v_i$:} A positive integer number, denoted by $L_{v_i}$, assigned to the node that captures length of the longest path in terms of the node count from any leaf node to $v_i$.  \\
\textit{Depth of a graph:} For a graph G=(V,E), the depth is defined as the largest $L_{v_i}$ for all nodes $v_i$ in V. \\
\textit{And Inverter Graph (AIG)}: A graph G=(V,E) where each vertex in V has in-degree of two; these vertices have conjunctive functionalities (2-input AND gates). Negative sign (inversion) is modeled as an attribute on edges of the graph.  
%%%%%%%%%%%%%%%%%%%%%%%%%%%%%%%%%%%%%%%%%%%%%%%%%%%%%%%%%%%%%%%%%%%%%%%%%%%%%%%%%%%%%
\subsection{ Q-learning Algorithm}
\label{RL_subsec}
Q-learning is a model-free, off-policy RL algorithm that maximizes the long-term reward; it involves a Q-agent, a set of states, and a set of actions that can be taken at each particular state. The Q-learning algorithm can be modeled as a function that assigns a real number to each (state, action) pairs: $Q:S \times A \rightarrow \Bbb R$. In the Q-learning algorithm at any time stamp \emph{t}, the Q-agent is at state $s_t$; by taking action $a_t$ it will enter a new state $s_{t+1}$. In this process, a local reward of $r(s_t,a_t)$ will be granted to the agent and the Q-value of the corresponding $(s_t,a_t)$ pair will be updated using the well-known Bellman-Ford equation below:

{\small
\begin{multline}
\label{Q_update}
Q(s_t,a_t) \leftarrow (1 - \alpha) \times Q(s_{t-1},a_{t-1}) +  \\
\alpha \times \biggl(  r(s_t, a_t) + \gamma \times \underset{a'}{\operatorname{max}} ~Q(s_{t+1},a')   \biggr)
\end{multline}
}
%%%%%%%%%%%%%%%%%%%%%%%%%%%%%%%%%%%%%%%%%%%%%%%%%%%%%%%%%%%%%%%%%%%%%%%%%%%%%%%%%%%%%
where $\alpha$ is the learning rate, $\gamma$ is the discount factor for future rewards and $Q(s_{t},a_{t})$ and $Q(s_{t-1},a_{t-1})$ are new and old Q-values, respectively. In a more general case, $\alpha$ can have different values for each pairs of state and action. In this case, it will be denoted by $\alpha (s_t,a_t)$.
\subsection{Asynchronous Advantage Actor-Critic (A3C) Algorithm}
\label{A3C_subsec}
Asynchronous Advantage Actor-Critic (A3C) algorithm \cite{mnih2016asynchronous} is one of the state-of-the-art deep reinforcement learning algorithms and an improvement over the previous Deep Q-Networks \cite{mnih2013playing} and the Advantage Actor-Critic (A2C) ones. Unlike previous deep reinforcement learning algorithms, in A3C, there are multiple agents each of which with its own network parameters and a copy of the global neural network that is on-board. These agents learn asynchronously by independently interacting with the environment and then they share the learned knowledge by updating the global network. A3C also combines the benefits of value-iteration and policy-gradient methods. As its name implies, A3C uses an advantage function, defined below, to learn how much a reward is better than the anticipated value instead of only learning the sign of the reward, which is an improvement over the policy-gradient method.     
{\small
\begin{equation}
    A(s,a) = Q(s,a) - V(s)
\end{equation}
}
where \emph{Q(s,a)} is the Q-value for taking action \emph{a} being at state \emph{s}, and \emph{V(s)} is the average value for the given state \emph{s}.
%%%%%%%%%%%%%%%%%%%%%%%%%%%%%%%%%%%%%%%%%%%%%%%%%%%%%%%%%%%%%%%%%%%%%%%%%%%%%%
\subsection{Cut-based AIG-Rewriting Algorithm}
\label{rw_subsec}
Cut-based AIG-rewriting \cite{mishchenko2006dag} is a greedy algorithm to reduce the size of an AIG graph by iterative selection of some sub-graphs rooted at each node in the AIG and trying different rewritten versions of the function at the root node; a rewritten version that reduces the node count more than others will be selected and will replace the current sub-graph. In this process, the functionality of the root node will always be preserved. For computing the said sub-graphs, cuts with up to four inputs (4-feasible cuts) are used. For an accurate definition of \emph{k-feasible cuts} see \cite{cong1994flowmap}. 

%%%%%%%%%%%%%%%%%%%%%%%%%%%%%%%%%%%%%%%%%%%%%%%%%%%%%%%%%%%%%%%%%%%%%%%%%%%%%%%%%%%%%
\section{Related Work}
\label{prior_work_sub_sec}
Hosny \emph{et al.} presented DRiLLS \cite{hosny2020drills} a deep reinforcement learning-based logic synthesis approach using the A2C algorithm. DRiLLS addresses the problem of large design space exploration and finds the best sequences for applying different optimization functions among a few given candidates; experimenting on some open source benchmark circuits, the authors reported an average area reduction of 13\%. Zhu \emph{et al.} \cite{zhu2020exploring} proposed a Markov Decision Process (MDP) formulation for logic synthesis and an RL algorithm that incorporates the Graph Convolutional Networks (GCNs) to explore the solution space; they reported improvements on different open source benchmark circuits compared to a classical heuristic-based approach. Yu \emph{et al.} \cite{yu2018developing} proposed an exact logic synthesis flow using a Convolutional Neural Network (CNN) to eliminate the need for human experts from the whole synthesis process; the authors argued that they could generate the best designs by the time for three large open source benchmark circuits, improving over the state-of-the-art logic synthesis tools.

In \cite{pasandi2020deep}, a deep learning-based approximate logic synthesis framework called \emph{Deep-PowerX} is presented to minimize the total switching power of digital circuits. Deep-PowerX operates on post-mapped netlists and identifies critical power nodes and replaces them with simpler nodes or completely removes them from the given netlist while consulting with an on-board pre-trained Deep Neural Network (DNN) to adhere to a given maximum error rate. Pasandi \emph{et al.}, presented Q-ALS: a reinforcement learning-based approximate logic synthesis engine. Q-ALS formulates the approximate logic synthesis problem as a Q-learning problem and gives solid definitions and formulations for the required state, action, and reward function. They reported significant improvements on area and delay compared to previous state-of-the-art approaches.    
%%%%%%%%%%%%%%%%%%%%%%%%%%%%%%%%%%%%%%%%%%%%%%%%%%%%%%%%%%%%%%%%%%%%%%%%%%%%%%%%%%%%%
%\section{AISYN: Our proposed framework}
\section{Methodology}
\label{ours_meth_sec}
As mentioned before, many optimization problems in logic synthesis are NP-Complete with no available optimal solutions. This is why many classical logic synthesis optimization functions follow a greedy approach, i.e., minimizing cost or maximizing saving at each step without having an idea of the global effect of this local minimization. The heuristical/greedy approaches can potentially lead to getting trapped into local minima which hinder optimizing the input circuit as much as needed. Our solution for this issue is to add AI-awareness to these optimization functions. More specifically, we have used two reinforcement learning algorithms, Q-learning and A3C to guide a logic synthesis optimizer to escape from local minima. For these RL algorithms to be effective, the problem formulation has to be done carefully; it is critical to have solid definitions for state, action, and to have an effective reward function that helps the RL-agent to learn efficiently by interacting with the environment (circuit). Otherwise, there will be no learning and therefore, no further improvements in QoR will be achieved. In this section, it will be shown how we define a logic synthesis optimization problem as an RL problem with state, action, reward function definitions and formulations with a fast training process. 

As for the logic synthesis optimization algorithgm, we identified a cut-based rewriting optimization function, part of technology independent optimization, as a key step that can have a major effect on the final QoR when it is guided by an effective RL algorithm. During a rewrite operation, the synthesis tool analyzes all possible alternative implementations of every cut and selects the best ones for maximizing the long term reward, translating into minimizing the cost function. We developed a system which can make series of random decisions (selection of cuts) for every node in the circuit during the rewrite stage, and eventually learn the quality of each of these decisions by back-propagating the final QoR metric of the design at the end of the synthesis process. It is a ``live reinforcement'' system, which means the randomness of a decision becomes more guided as it continuously learns the quality of that decision once one round of optimization is finished. In the following, we will explain how we formulate the cut-based rewriting optimization function as an RL problem. 
%%%%%%%%%%%%%%%%%%%%%%%%%%%%%%%%%%%%%%%%%%%%%%%%%%%%%%%%%%%%%%%%%%%%%%%%%%%%%%%%%%%%%%%%%%%%%%%%%%%%%
\subsection{Modified AIG-rewriting function}
\label{mod-rewriting_sub_sec}
\begin{figure}[t]
\centering
\includegraphics[width=0.35\textwidth]{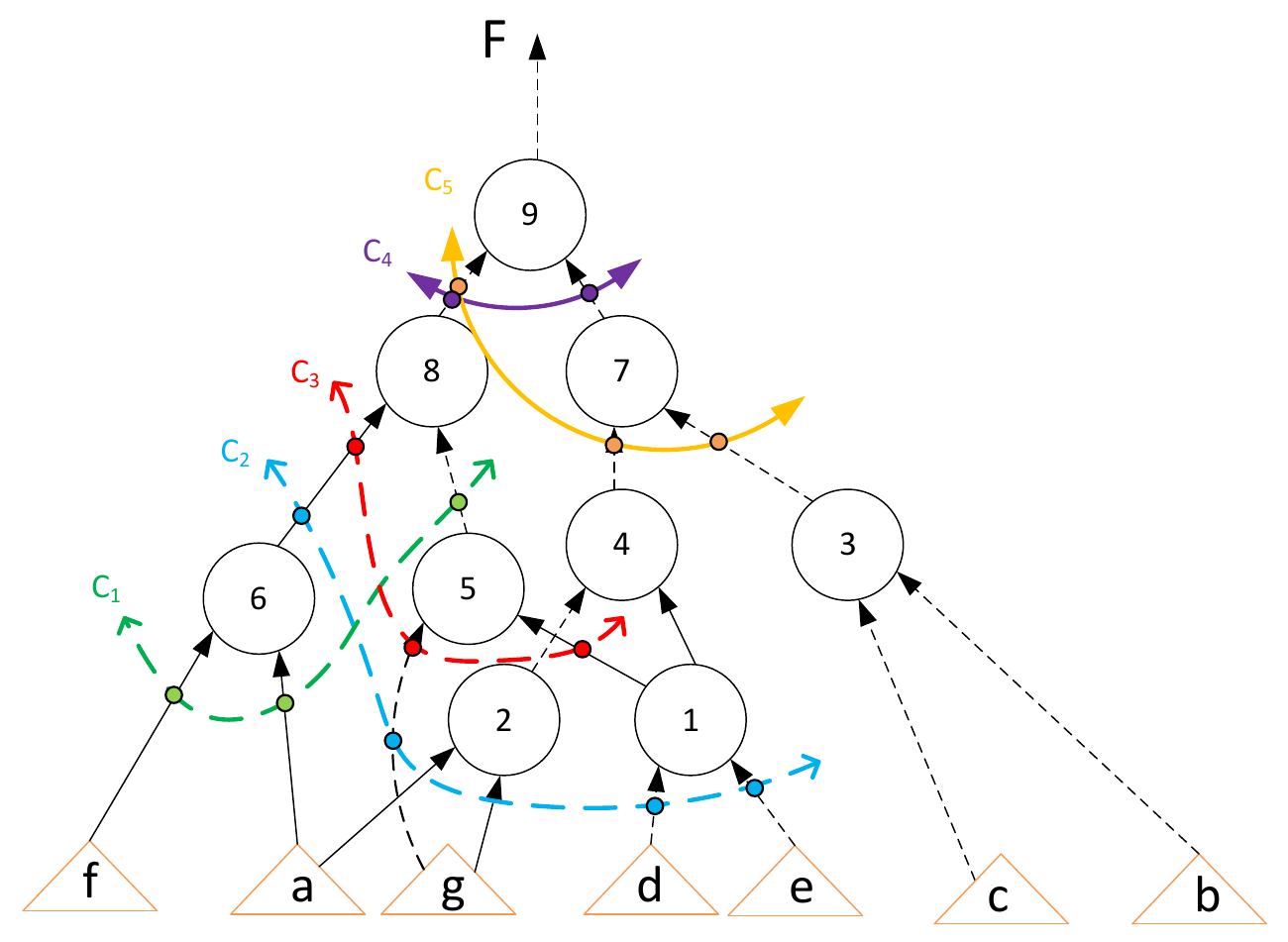}
\caption{A test AIG with nine nodes. Some of the 4-feasible cuts for nodes 8 and 9 are shown: $C_4$ and $C_5$ for node 9, and $C_1$, $C_2$, and $C_3$ for node 8.}
\label{mod_rw_fig}
\end{figure}

Before going into details of our RL-based rewriting formulations and algorithms, we need to present a modified AIG-rewriting function that is flexible enough for the RL-based setup. The modified rewriting function should make it possible to select any cut for a node among its k-feasible cuts. This allows us to explore and take random actions rather than sticking to certain cuts e.g. cuts that give the highest local gain as is in the classical rewriting function. An example AIG is shown in Fig. \ref{mod_rw_fig}. Some of the 4-feasible cuts of nodes 8 and 9 are shown in this figure. Suppose rewriting functions of nodes 8 and 9 based on inputs of cuts $C_3$ and $C_5$, respectively yield the best gains for rewriting functions of these nodes. In this case, the classical rewriting algorithm will always choose $C_3$ for node 8 and $C_5$ for node 9. In our RL-based rewriting process, we need to be able to select either of $C_1$, $C_2$, and $C_3$ cuts for node 8 and either of $C_4$ and $C_5$ for node 9, with any combination of the two. Our modified rewriting function provides this luxury for us. Below, it will be explained how choosing each of these cuts will correspond to taking an action in the RL terminology, making the exploration phase possible in the RL setup and hence helping us performing the training phase in the RL-based logic rewriting process. We have studied usage of two RL algorithms in this paper, Q-learning and A3C, which are explained below. 
%%%%%%%%%%%%%%%%%%%%%%%%%%%%%%%%%%%%%%%%%%%%%%%%%%%%%%%%%%%%%%%%%%%%%%%%%%%%%%%%%%%%%
%%%%%%%%%%%%%%%%%%%%%%%%%%%%%%%%%%%%%%%%%%%%%%%%%%%%%%%%%%%%%%%%%%%%%%%%%%%%%%%%%%%%%
\subsection{Q-learning-based rewriting algorithm}
\label{Q-learning_rewrite_subsec}
In the rewriting process, the cut-based AIG-rewriting algorithm as explained in Section \ref{rw_subsec} computes up to \emph{p} different cuts for each node of the given AIG. Having $n$ nodes in the graph, this suggests an upper bound of $p^n$ for the state space, the total number of possible AIGs that can be achieved through AIG-rewriting. Selecting any of these cuts at each time stamp \emph{t} and updating the graph using the resulting rewritten local sub-graph will take the system from state \emph{x} to state \emph{y}, with a probability $P_a(x,y)$ where \emph{a} is the cut selection action. We have limited the number of actions per state and in our setup it is clear that the next state \emph{y} will be only dependent on the current state \emph{x}, satisfying the memory-less property of Markov chains. Therefore, the presented AIG-rewriting can be modeled as a MDP, hence, Q-learning can be used to solve it optimally. There are four parameters in MDP that should be carefully defined: $\chi$: the (finite) state space, $A$: the (finite) action space, $P$: the transition probabilities, and $r$: the reward function. In the Q-learning-based rewriting algorithm, these parameters are defined as below:
\begin{itemize}
    \item $\chi$:  different AIG representations of a given circuit. 
    \item $A$: different k-feasible cuts that can be selected for nodes of the AIG.
    \item $P$: is always 1, because in our problem selecting any action will definitely take the system to the determined next state.
    \item $r$: gain (e.g. improvements on AIG node count) achieved by performing a cut rewriting operation on an AIG node.
\end{itemize}
\begin{theorem}
\label{Q_theorem}
The presented AIG-rewriting problem can be solved optimally using the Q-learning algorithm. 
\end{theorem}
\textit{Proof}: In \cite{melo2001convergence}, it is proven that a finite $MDP (\chi, A, P,r)$ can be solved optimally (converging to optimal Q-function) using Q-learning with an update rule as in Eq. \ref{Q_update} while $0 \leq \alpha \leq 1$ is dependent on both state and action ($\alpha(s_t,a_t)$) and satisfies the following requirements for all state-action pairs:
\begin{equation}
\label{a_eq1}
    \sum_{t} \alpha(s_t,a_t) = \infty  
\end{equation}
\begin{equation}
\label{a_eq2}
    \sum_{t} \alpha^2(s_t,a_t) < \infty  
\end{equation}
As in \cite{melo2001convergence}, if we visit each state-action pair infinitely enough, choosing $\alpha(s,a)$=$1/k$, will satisfy the the above requirements, where \emph{k} is the number of times the $(s_t,a_t)$ pair is visited. It is easy to show that $\sum_{k=1}^{\infty} \frac{1}{k}$ is lower  bounded by $\int_{0}^{\infty} \frac{1}{k}dk$=$ln(\infty)$=$\infty$. Therefore, Eq. \ref{a_eq1} is satisfied. Similarly, it can be shown that $\sum_{k=1}^{\infty} \frac{1}{k^2}$ is upper bounded by $\int_{0}^{\infty} \frac{1}{k^2}dk$=1, therefore, Eq. \ref{a_eq2} is also satisfied.  $\blacksquare$

However theorem \ref{Q_theorem} proves that Q-learning can converge to optimal solution, but since it requires visiting each state-action pair infinite times, it is not guaranteed that in practice it can yield the optimal solution. Even trying all state-action pairs will require huge resources; for a small 4-bit adder, we have $p=9$ different cuts per node, and $n=48$ total nodes that makes the state space upper bounded by $9^{48} \approx 6.4 \times 10^{45}$. A more accurate state space for this adder is $4.4 \times 10^{15}$ (note that many nodes will have much fewer number of cuts compared to the maximum value of 9). This is a very large state space even for a small circuit, hence, it is not very practical to have these many states in an RL formulation for the rewriting problem\footnote{In this calculation we assumed that the root node function for each cut has only one rewritten version, which normally is much more than one, therefore, a more realistic value for the number of states is even larger than these values.}. This is why in our Q-learning-based rewriting algorithm, we resort to using an approximate representation for the state of the system.    

In our Q-learning-based rewriting algorithm (\emph{Q-drw} from here after), similar to \cite{pasandi2019approximate}, nodes are used to represent approximate states. A Q-agent will traverse the AIG form of the input circuit in a topological order and at each step, it will go from node $v_i$ to $v_j$, taking the state of the system from $i$ to $j$. When the Q-agent is at node $v_i$, an action that it can take is selecting a cut among k-feasible cuts of node $v_i$ for rewriting the function at the root node $v_i$. Selecting any of these cuts and trying different rewritten versions of the function of the root node based on inputs of the cut will take the system to a new unique state. However, using the said approximate model the Q-agent will always end up at node $v_j$ when it takes an action being at node $v_i$, therefore, its state will always go from $i$ to $j$. The difference between taking different actions at node $v_i$ however will appear in the quality of final results and therefore will be compensated by a different reward value. Since the goal of the agent is to maximize the long-term reward, therefore, the Q-agent will effectively learn which cuts are the best to choose at each node.  
%%%%%%%%%%%%%%%%%%%%%%%%%%%%%%%%%%%%%%%%%%%%%%%%%%%%%%%%%%%%%%%%%%%%%%%%%%%%%%%%%%%%%
\begin{figure}[t]
\centering
\includegraphics[width=0.45\textwidth]{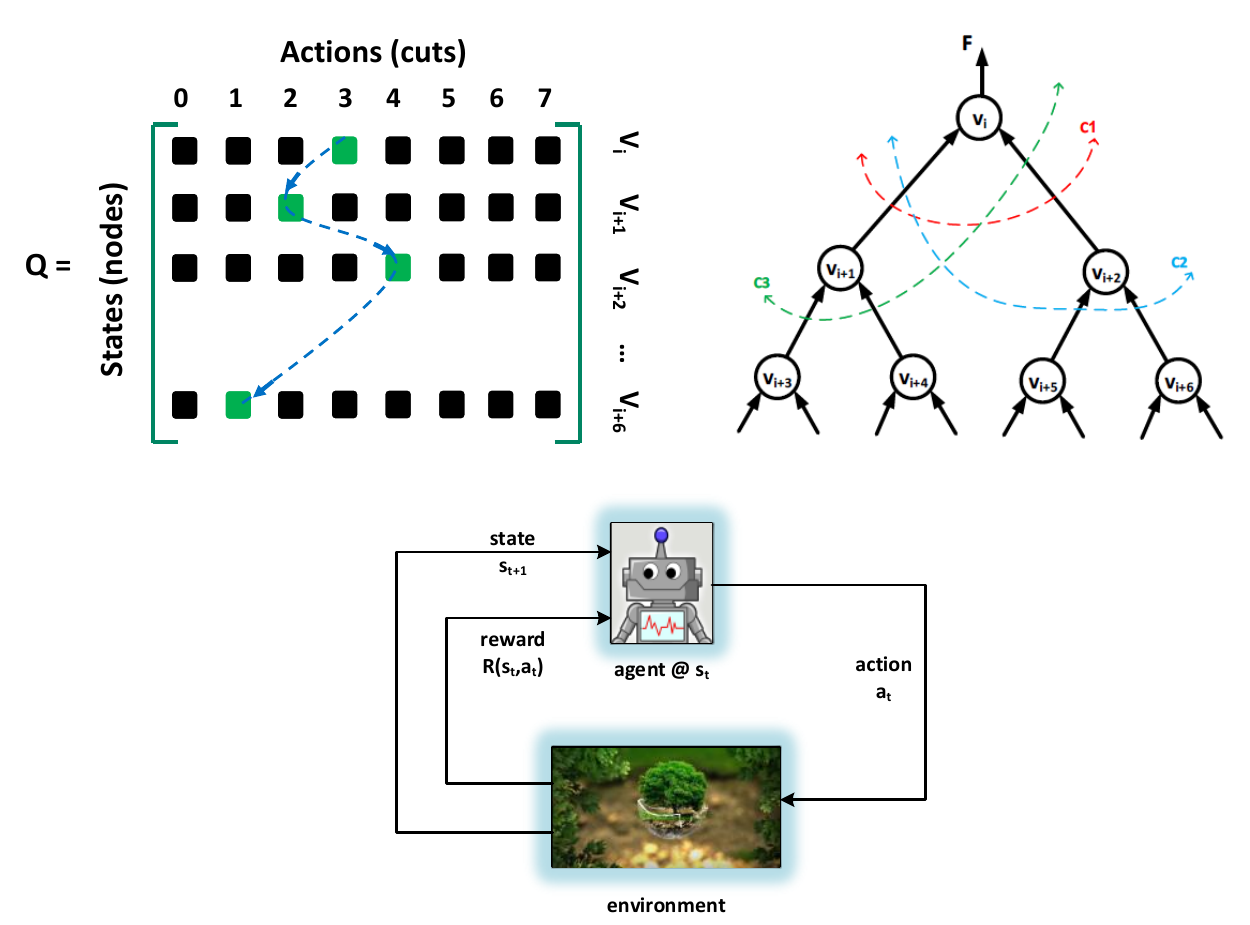}
\caption{An example for our Q-learning-based rewriting algorithm. A Q\_matrix with having a row for each node of the given circuit, representing an approximate state, and a column for each action is constructed. The shown AIG has seven nodes, hence, seven rows for the Q\_matrix, and there are a maximum of eight 4-feasible cuts, hence, eight columns for the matrix.}
\label{Q-rw_frame_fig}
\end{figure}  
%%%%%%%%%%%%%%%%%%%%%%%%%%%%%%%%%%%%%%%%%%%%%%%%%%%%%%%%%%%%%%%%%%%%%%%%%%%%%%%%%%%%%
The Q-agent will receive two types of rewards: 
\begin{itemize}
    \item \textit{local reward}: A small reward corresponding to a local gain that is achieved from taking an action. 
    \item \textit{global reward}: A large terminal reward that is compensated at the end of an episode based on the quality of a series of actions.  
\end{itemize}
Selecting a cut $C_j$ for node $v_i$ and performing the local rewriting step will result in having a new AIG for the sub-graph rooted at $v_i$ with fewer or more (by $\Delta v$ amount) nodes compared to the old one. If the number of nodes is fewer, then the agent will receive a positive local reward of $+\Delta v \times K_1$. Otherwise, it will get a negative reward of $-\Delta v \times K_2$, where $K_1$ and $K_2$ are two normalization factors. Once a full node traversal is done (one episode), the rewritten AIG will be passed to a technology mapper to bind logic gates to its nodes. The cost (delay, area, power) of the new circuit will be higher or lower (by $\Delta C$ amount) than the baseline. If the cost is lower, the agent will get a positive terminal reward of $+\Delta C \times K'_1$. Otherwise, it will get a negative reward of $-\Delta C \times K'_2$, where $K'_1$ and $K'_2$ are two normalization factors.

Our experiments show that if magnitudes of global rewards are much larger than the local ones, it will help the Q-agent to learn better and converge faster. For this purpose, we always choose values for the normalization factors such that they satisfy the following: $K'_1 >> K_1$ and $K'_2 >> K_2$.
In the training phase, an $\epsilon$-greedy \cite{mnih2013playing} approach is followed, which means that an action with the maximum Q-value with probability 1-$\epsilon$ and a random action with probability $\epsilon$ will be selected by the Q-agent. We start with larger values (closer to 1) for $\epsilon$ and gradually cool it down. This will gradually move the Q-agent from exploration phase (taking more random actions in the beginning) to the exploitation phase (taking actions based on the gained knowledge).  

Fig. \ref{Q-rw_frame_fig} shows an example of our table-based implementation for Q-learning-based rewriting algorithm. There are as many rows in the Q\_matrix as the number of nodes in the AIG form of the given circuit (seven in this example). Also, since 4-feasible cuts are used, the maximum number of 4-cuts for a node in the graph determines the number of columns in the Q\_matrix (eight in this example). Before training, the Q\_matrix is initialized to random values between a \emph{min} value and a \emph{max} value.

Algorithm \ref{Q-rw:training} shows the pseudo code for our Q-learning-based rewriting algorithm. Inputs of the algorithm are the input circuit and a technology library. In lines 1-6 hyper-parameters are initialized and the Q\_matrix is constructed. The user also has the option to set these initial values through command line arguments. In lines 7-8, the given circuit is converted to AIG and the number of states, which is equal to the node count of this AIG, is calculated. In lines 9-21 the main training process is being done. More specifically, the \emph{for loop} in lines 11-16 determines the set of actions that will be taken in an episode based on an $\epsilon$-greedy approach. Finally, in lines 22-23, the optimized netlist will be obtained and returned.  
%%%%%%%%%%%%%%%%%%%%%%%%%%%%%%%%%%%%%%%%%%%%%%%%%%%%%%%%%%%%%%%%%%%%%%%%%%%%%%%%%%%%%
\begin{algorithm} [t]
{\small
\caption{Q-learning-based rewrite}\label{Q-rw:training}
\DontPrintSemicolon % Some LaTeX compilers require you to use \dontprintsemicolon instead
\KwIn{Input circuit: $N_{in}=(V_{in},E_{in})$\\ technology library: $\mathfrak{L}$ } 
\KwOut{An optimized netlist: $\mathbb{N_{OPT}}$=$\mathbb{(V_{OPT}, E_{OPT})}$}
{Initialize: \\  
\tab Number of Episodes: $Episodes$ \\
\tab Learning rate: $\alpha$ \\
\tab Discount factor: $\gamma$ \\
\tab Exploration parameter: $\epsilon$ \\
Construct and randomly fill in the Q\_matrix \\}
$N_{AIG}=(V_{AIG},E_{AIG}) \leftarrow$ Convert $N_{in}$ to AIG format. \\
$num\_states$ = n($V_{AIG}$). \\
%$\epsilon$=1.0 \\
\For{itr in range($Episodes$)}{
    \emph{actions} = list()  \\
    \For{idx in range($num\_states$)}{

        \If{random.uniform(0,1) $ < \epsilon$}{
            act = random.randrange(0,len(Q\_matrix[idx]))
        }
        \Else{
            act = argmax(Q\_matrix[idx])
        }
        \emph{actions}.append(act)
    }
    Perform a full rewrite using \emph{actions}.\\
    Perform tech. mapping and calculate cost function. \\
    Calculate \emph{reward} and update Q\_matrix. \\
    Keep track of \emph{best scores}. \\
    Decrease $\epsilon$.
}
$\mathbb{N_{OPT}}$ = get the best netlist based on learnt knowledge.\\

\Return{$\mathbb{N_{OPT}}$}\;
}
\end{algorithm}
%%%%%%%%%%%%%%%%%%%%%%%%%%%%%%%%%%%%%%%%%%%%%%%%%%%%%%%%%%%%%%%%%%%%%%%%%%%%%%%%%%%%%
\begin{figure}[t]
\centering
\includegraphics[width=0.45\textwidth]{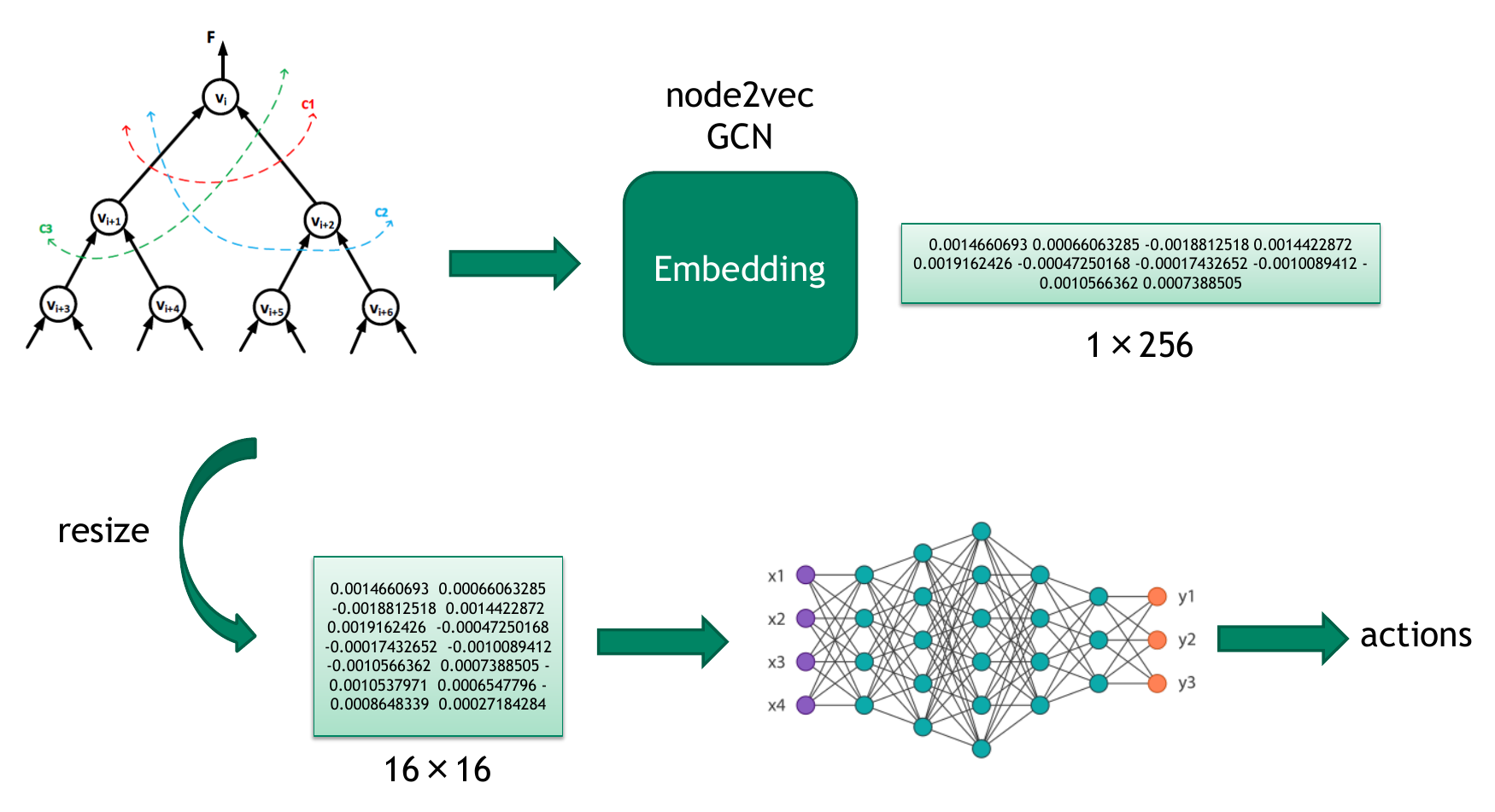}
\caption{Generating node embeddings and feeding them to neural networks in A3C-drw framework. The AIG representation of the input circuit is given to a node embedder to generate feature vectors for its nodes. This feature vector is then resized and fed into the on-board deep neural network in A3C framework to predict actions (cuts) being at each state (node). Note that the node embedding is done only once for each AIG and then it is used multiple times by different agents in A3C-drw framework.}
\label{A3C-rw_frame_fig}
\end{figure}  
%%%%%%%%%%%%%%%%%%%%%%%%%%%%%%%%%%%%%%%%%%%%%%%%%%%%%%%%%%%%%%%%%%%%%%%%%%%%%%%%%%%%%
\subsection{A3C-based rewriting algorithm}
\label{A3C_rewrite_subsec}
Similar to the table-based Q-drw, in A3C-based rewriting algorithm (A3C-drw), nodes are used to represent approximate states of the system. However, in A3C-drw, to be able to feed a state into on-board neural networks, a more general node representation is needed. For this purpose and to generate node embeddings, we have tried different representation learning methods including: GCN \cite{kipf2016semi} and node2vec \cite{grover2016node2vec}. Fig. \ref{A3C-rw_frame_fig} shows how node embeddings are generated and fed into on-board neural networks in the A3C-drw framework: First, an AIG representation of the given circuit is passed to a node embedding engine to generate feature vectors for all nodes. These feature vectors are resized to have a standard square shape (2D array) and fed into the neural networks. Then, these networks predict the best actions to be taken being at any node of the input AIG. In the training phase, these predicted actions will result in some positive or negative rewards which will be back-propagated to update weights and biases values of the neural networks. We have used deep neural networks for both actor and critic, with the following architecture: there is a convolution layer, followed by a max pooling layer and a dropout layer. Then there are two fully connected layers each followed by a dropout layer.

Similar to Q-drw, in A3C-drw, an action that can be taken at node $n_i$ is defined as selecting a cut out of 4-feasible cuts of the node $n_i$. A3C-drw uses a similar reward function that is mentioned in Section \ref{Q-learning_rewrite_subsec} for Q-drw. 
%%%%%%%%%%%%%%%%%%%%%%%%%%%%%%%%%%%%%%%%%%%%%%%%%%%%%%%%%%%%%%%%%%%%%%%%%%%%%%%%%%%%%
\subsection{Tuning hyper-parameters}
\label{hyper_sub_sec}
Similar to any other AI system, in our AI-driven logic synthesis framework, having the right values for hyper-parameters and thus, tuning them for this purpose is very important. For example, our experiments show that varying normalization factors presented in Section \ref{Q-learning_rewrite_subsec} for calculating the reward that is given to the agent plays a significant role in convergence of the system as well as in the quality of results. We have used $K_1$=$K_2$=1 for normalization factors in local reward calculations. For the normalization factors that are used in global reward calculations, we have four sets of numbers: \emph{small}, \emph{medium}, \emph{large}, and \emph{supersize}; these are utilized based on the following scenario: if taking an action improves the cost function over the baseline, a \emph{large} value will be used for $K'_1$; if it improves over baseline as well as the best achieved value so far, the RL-agent is rewarded significantly by setting $K'_1$ to a \emph{supersize} value. The best value for the cost function is initially set to the same value as the baseline, but if a better result is found, this best value will be updated. There is a $3^{rd}$ case: for circuits that are harder to optimize and therefore very difficult to get better results than the baseline, the RL-agent will not be granted much positive rewards in the beginning episodes based on what is explained so far. This can result in the agent being lost and not being able to learn anything, because the agent won't have any idea of which actions are good. To solve this issue, we keep track of the best score that is achieved so far (it might be worse than the base score in the beginning) and if the RL-agent can improve on that, a positive reward by using a \emph{medium} value for $K'_1$ is given to the agent.  
$K'_2$ always has a \emph{small} value. In this paper, we have used the following numbers: 10 for \emph{small}, 30 for \emph{medium}, 50 for \emph{large}, and 200 for \emph{supersize}. 
%%%%%%%%%%%%%%%%%%%%%%%%%%%%%%%%%%%%%%%%%%%%%%%%%%%%%%%%%%%%%%%%%%%%%%%%%%%%%%%%%%%%%
\begin{figure*}[t]
        \centering
        \begin{subfigure}[!t]{0.25\textwidth}
                \centering
                \includegraphics[width=\textwidth]{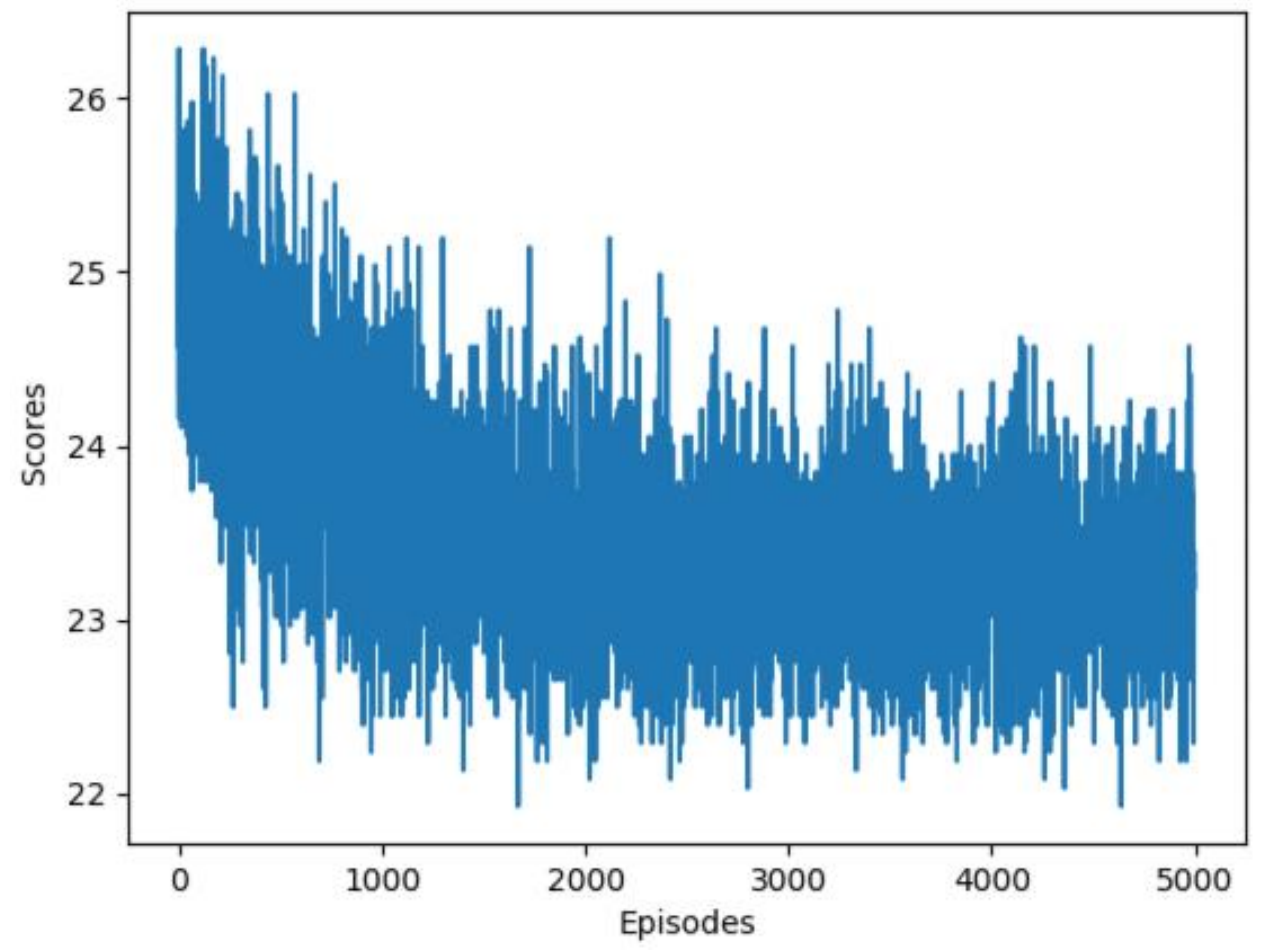}
                \caption{}
                \label{scores_fig}
        \end{subfigure}
        %%%%%%%%%%%%%%%%%%%%%%%%%%
        \begin{subfigure}[!t]{0.25\textwidth}
                \centering
                \includegraphics[width=\textwidth]{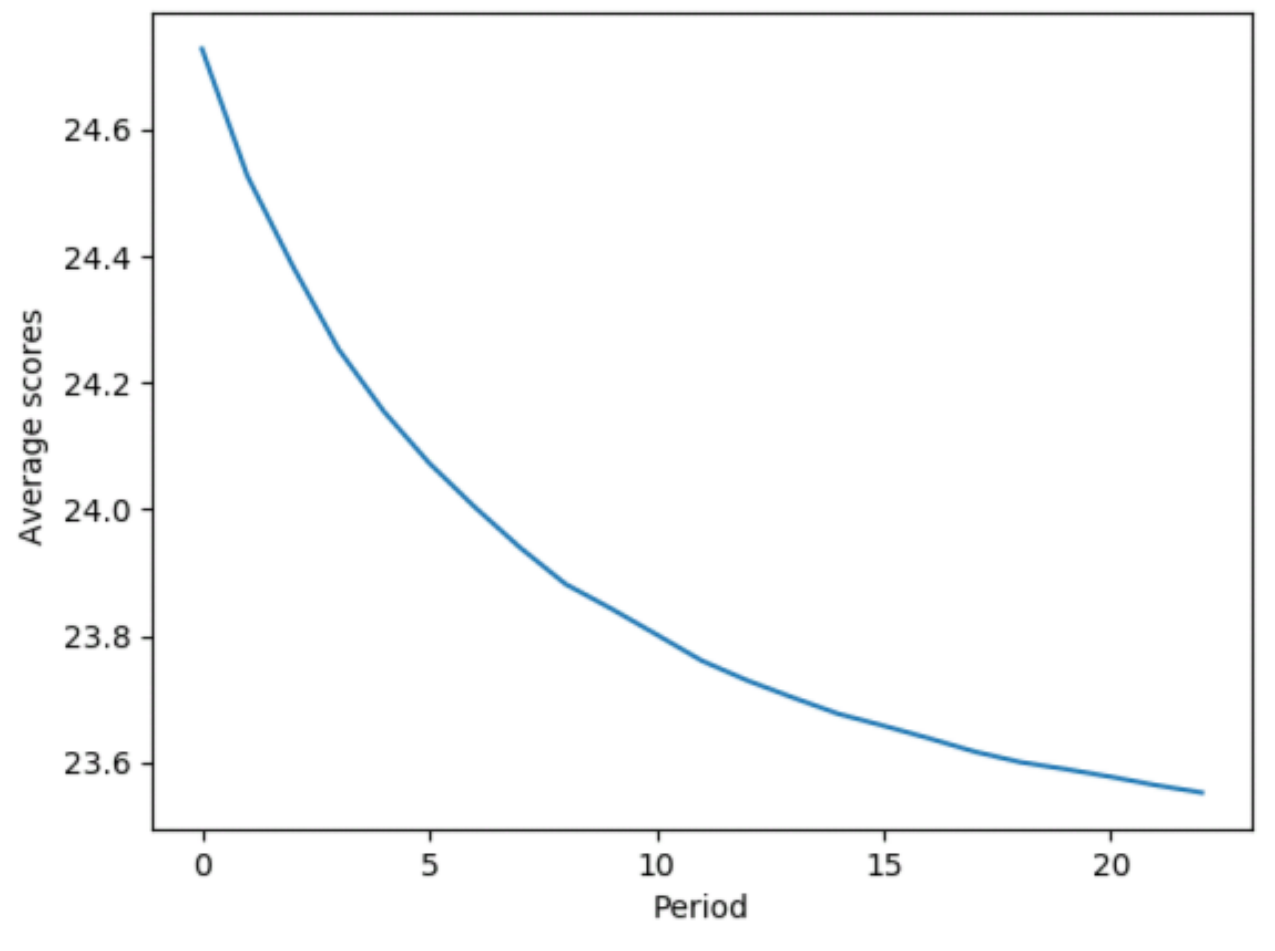}
                \caption{}
                \label{avg_scores_fig}
        \end{subfigure}
        %%%%%%%%%%%%%%%%%%%%%%%%%%
        \begin{subfigure}[!t]{0.25\textwidth}
                \centering
                \includegraphics[width=\textwidth]{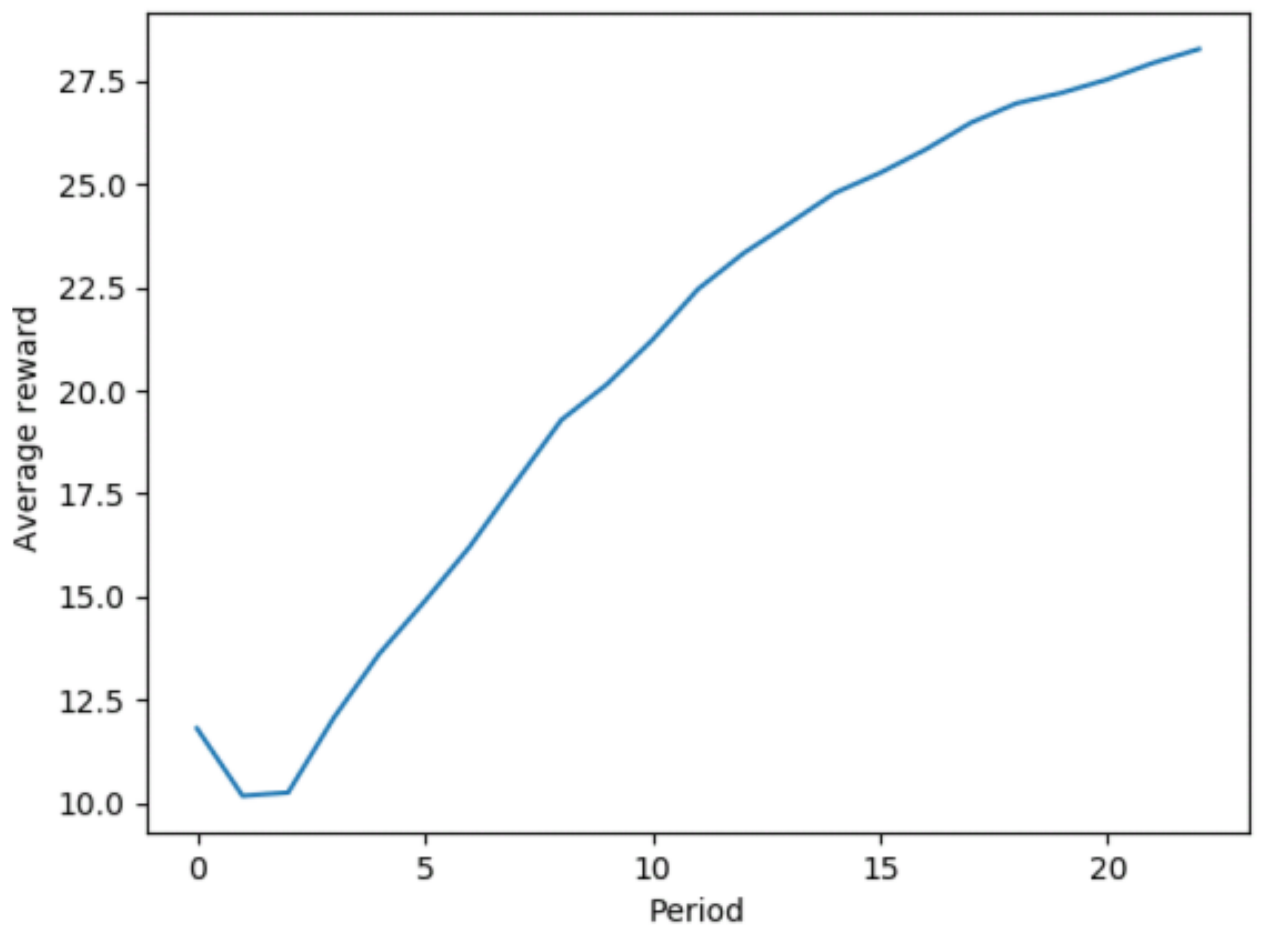}
                \caption{}
                \label{rewards_fig}
        \end{subfigure}
        \caption{Learning curves for a small combinational circuit (c432 of ISCAS) for 5000 episodes. (a) score per episode, (b) average scores per period (=200 episodes), (c) average rewards given to the RL-agent per period. As seen, the average scores are decreasing as the agent moves forward; since the score here is the cost function, this means that the solutions that the RL-agent creates are getting better as it goes. As seen in the right most graph, for the first few hundred episodes, the average reward is decreasing. This is because the agent is in the exploration phase and tends to take many bad actions; this changes after a while when the agent got a better understanding of the environment, hence, taking better actions and receiving more positive rewards. }
        \label{scores_rewards_fig}
\end{figure*}
%%%%%%%%%%%%%%%%%%%%%%%%%%%%%%%%%%%%%%%%%%%%%%%%%%%%%%%%%%%%%%%%%%%%%%%%%%%%%%%%%%%%%

Fig. \ref{scores_rewards_fig} shows graphs for scores per episode, their averages over different periods (=200 episodes), and the average reward on each period. These graphs are obtained for a small combinational circuit (ISCAS c432), which made it easy to experiment for a larger number of episodes. An episode is defined as a full traversal of the whole AIG representation of the given circuit by the RL-agent while taking different actions at each time stamp. From the shown graphs, it is evident that as the agent moves forward (later episodes), the quality of rewritten AIGs are getting better, therefore, the cost is getting decreased. This shows an organic learning process which is a clear indication of a successful AI-driven rewriting framework. Fig. \ref{rewards_fig} shows that for the first few hundreds of episodes, the average reward that the agent receives is decreasing. This makes sense, because in the beginning episodes while the agent is in the exploration phase, it is not familiar with the environment yet and tends to take many bad actions. However, after around 500 episodes, the agent has learned well enough such that it can take better actions, resulting in an increase in the average reward that is granted to it. This is why the graph starts increasing and continues the trend for the rest of the training process. 
%%%%%%%%%%%%%%%%%%%%%%%%%%%%%%%%%%%%%%%%%%%%%%%%%%%%%%%%%%%%%%%%%%%%%%%%%%%%%%%%%%%%%
\section{Experimental Results}
\label{exper_sec}
We have used Python v3.6.6, Tensorflow v1.13.1, Keras v2.2.4, and Gym v0.17.2 for implementing the AI related algorithms used in this paper. For A3C, we started with a code \cite{A3C_code} written for CartPole game and modified it to meet our needs. Also, ABC \cite{synthesis2011abc}, UC Berkeley's open source logic synthesis tool is used to extract experimental results and also to implement our modified version of the rewriting function. The rewriting function that is used for extracting baseline results and also implementing our modified rewriting function is \emph{drw} which is a newer and more stable version of \emph{rw} command. We experimented on both industrial and open source benchmark circuits and achieved great results on both for our AI-driven logic synthesis framework. The results that are presented in this section are for classical combinational benchmark suite of ISCAS \cite{iscas} as well as arithmetic benchmarks of EPFL suite \cite{EPFL_bench}. We also include results for an important industrial block in this section. 

Two different reward functions are used: area for the case of experimenting on the open source benchmark circuits and total net toggle rate for the reported industrial block. 
We have also used a standard cell library with thousands of cells designed using TSMC's 16nm technology. A few baselines are used in this section including: ABC's \emph{map} command with no rewrite, ABC's \emph{drw} command plus \emph{map}, and 10 consecutive \emph{drw}s plus \emph{map}. We will be comparing results for \emph{Q-drw} plus \emph{map} and \emph{A3C-drw} plus \emph{map} with those of the baselines. 
%%%%%%%%%%%%%%%%%%%%%%%%%%%%%%%%%%%%%%%%%%%%%%%%%%%%%%%%%%%%%%%%%%%%%%%%%%%%%%%%%%%%%
\begin{table}[t]
\centering
\scriptsize
\caption{Total cell area post synthesis for five different methods: \emph{no drw}: no rewrite + map,  \emph{drw}: ABC's rewrite + map, \emph{Q-drw}: Q-learning-based rewrite + map, and \emph{A3C-drw}: A3C-based rewrite + map. }
\begin{tabular}{@{}cccccccc@{}}
\toprule
\textbf{circuit}  &	\textbf{no drw}  &	\textbf{drw}  & 	\textbf{10 drw}	&  \textbf{random} & \textbf{Q-drw}  & 	\textbf{A3C-drw} \\
\midrule
c1355	&114.57	&139.03	&139.03	&106.01	&82.11	&84.29  \\
\midrule
c1908	&91.7	&97.3	&96.99	&107.72	&79.37	&75.53  \\
\midrule
c2670	&163.56	&138.31	&136.44	&153.96	&127.11	&129.86  \\
\midrule
c3540	&235.04	&201.24	&201.24	&203.93	&178.69	&185.12  \\
\midrule
c432	&61.9	&45.15	&44.22	&48.21	&37.48	&39.19  \\
\midrule
c499	&102.75	&112.23	&112.23	&109.38	&79.32	&73.46  \\
\midrule
c5315	&235.41	&219.44	&223.85	&247.48	&228.56	&230.95  \\
\midrule
c6288	&377.19	&386.78	&377.55	&372.67	&348.94	&348.16  \\
\midrule
c7552	&348.26	&286.16	&275.17	&333.74	&303.99	&303.16  \\
\midrule
c880	&67.29	&53.08	&50.54	&55.08	&48.47	&50.7  \\
\midrule
sin	   &115.92	&106.12	&103.11	&108.22	&99.93	&100.98  \\
\midrule
max	   &457.38	&432.97	   &432.97	&430.94	   &416.74	&415.81  \\
\midrule
mult	&491.63	&437.37	&435.92	&477.45	&441.57	&439.38  \\
\midrule
log2	&495.51	&465.85	&466.93	&466.87	&449.98	&452.76  \\
\midrule
adder	&181.7	&181.7	&181.7	&187.55	&177.5	&178.12 \\
\bottomrule
average &235.98	&220.18	&218.52	&227.28	&206.65	&207.16 \\
\label{Table_area}
\end{tabular}
\end{table}
%%%%%%%%%%%%%%%%%%%%%%%%%%%%%%%%%%%%%%%%%%%%%%%%%%%%%%%%%%%%%%%%%%%%%%%%%%%%%%%%%%%%%
\begin{figure}[t]
\centering
\includegraphics[width=0.4\textwidth]{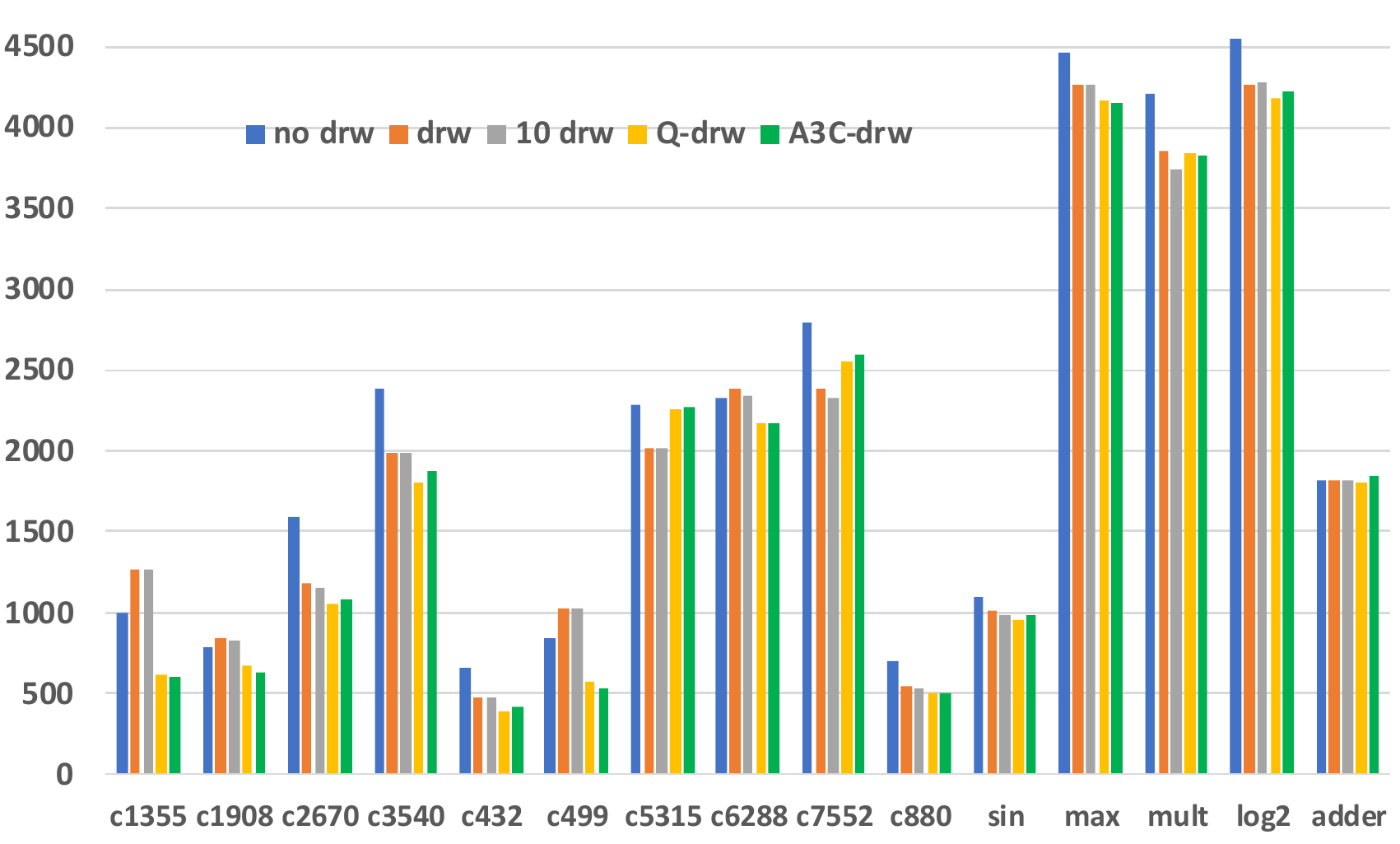}
\caption{Total edge count post synthesis for five methods presented in Section \ref{exper_sec}. For better exhibition purposes, data for \emph{log2}, \emph{mult}, and \emph{sin} is scaled down by a factor of 10. }
\label{Edge_exper_fig}
\end{figure}  
%%%%%%%%%%%%%%%%%%%%%%%%%%%%%%%%%%%%%%%%%%%%%%%%%%%%%%%%%%%%%%%%%%%%%%%%%%%%%%%%%%%%%
\begin{figure}[t]
\centering
\includegraphics[width=0.4\textwidth]{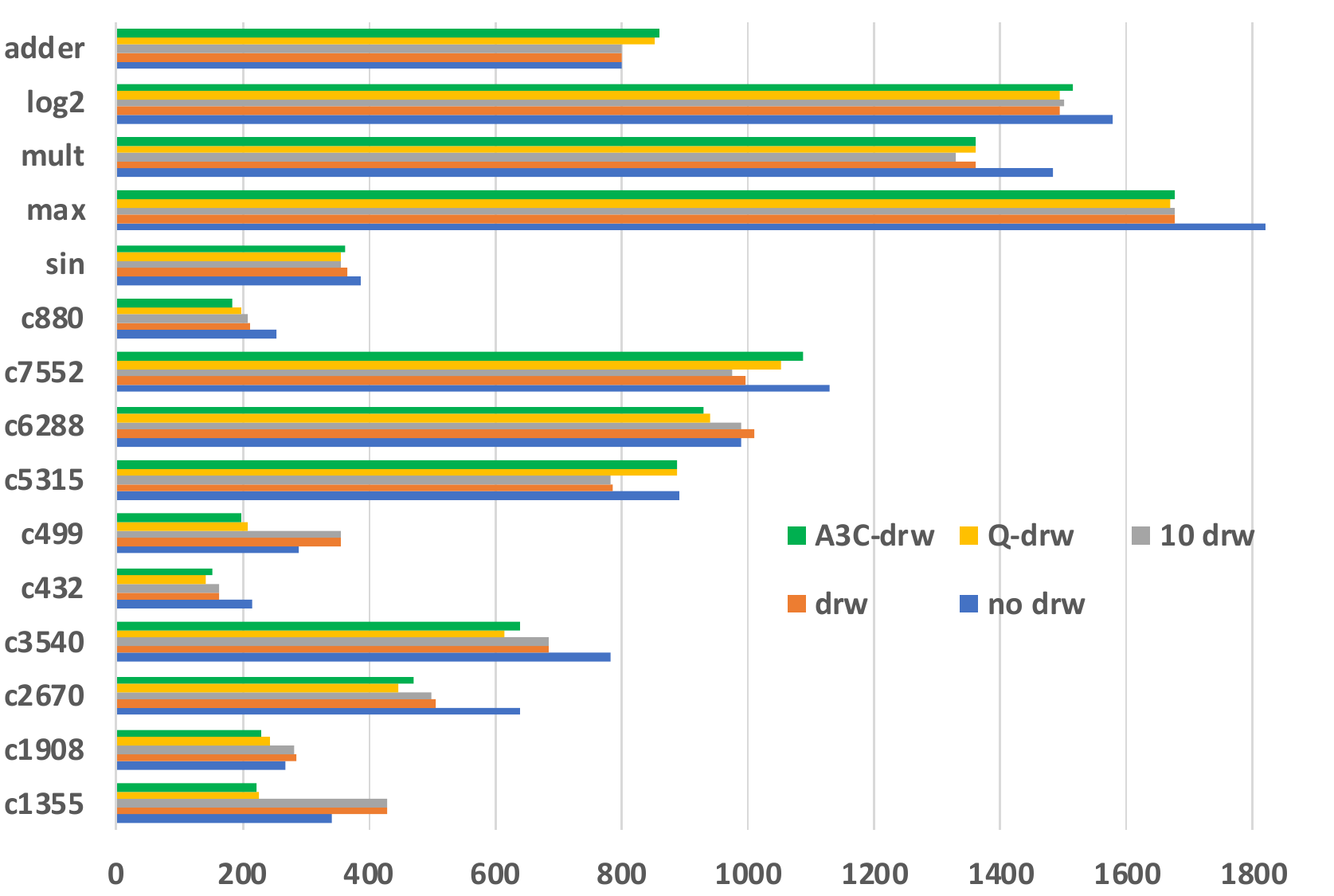}
\caption{Total node count post synthesis for five methods presented in Section \ref{exper_sec}. For better exhibition purposes, data for \emph{log2}, \emph{mult}, and \emph{sin} is scaled down by a factor of 10. }
\label{Node_exper_fig}
\end{figure}  
%%%%%%%%%%%%%%%%%%%%%%%%%%%%%%%%%%%%%%%%%%%%%%%%%%%%%%%%%%%%%%%%%%%%%%%%%%%%%%%%%%%%%

Tables \ref{Table_area}, \ref{Table_run-time} and Figs. \ref{Edge_exper_fig}, \ref{Node_exper_fig} show experimental results for total cell area, run-time values, total edge count, and total node count for five different methods: \emph{no drw}: no rewrite operation applied plus technology mapping,  \emph{drw}: ABC's rewrite plus map, \emph{10 drw}: 10 consecutive drw plus map, \emph{Q-drw}: Q-drw plus map, and \emph{A3C-drw}: A3C-drw plus map. \emph{Q-drw} could reduce the total area by up to 65.2\% for \emph{c432} circuit and 69.3\% for \emph{c1355} circuit compared to \emph{no drw} and \emph{drw}, respectively. Similarly, \emph{A3C-drw} could reduce the total area by 57.9\% for \emph{c432} and 64.9\% for \emph{c1355} when it is compared to \emph{no drw} and \emph{drw}, respectively. On average for 15 benchmark circuits, \emph{Q-drw} reduced the total area by 21.6\% and 13.4\% and \emph{A3C-drw} reduced it by 20.9\% and 13.1\% compared to \emph{no drw} and \emph{drw}, respectively. There is also a set of results in Table \ref{Table_area} corresponding to a random agent that selects random cuts during the rewriting process. As seen in this table, the area results for this random agent is significantly worse than both Q-drw and A3C-drw, which shows only taking actions different from the greedy approach of drw won't take us anywhere and there should be a reasonable learning setup for the agent to actually learn how to take good actions that result in some improvements. To remove effect of noise and any bias, the results presented in Table \ref{Table_area} for the random agent are average of 10 different tries.

Similar to area savings, significant savings are observed for total node count and total edge count in case of using Q-drw and A3C-drw compared to the baselines. For example, \emph{Q-drw} and \emph{A3C-drw} could reduce total node count by averages of 14.1\% and 14.5\% compared to \emph{drw}; the corresponding average savings for edge counts are 17.8\%, 18.3\%. 
These savings come at a small expense of 4.2\% and 0.6\% (on average) increase in critical path delay for circuits generated by \emph{Q-drw} compared to those generated by \emph{no drw} and \emph{drw}, respectively. The average delay degradation for \emph{A3C-drw} when it is compared to circuits generated by \emph{no drw} is 4.7\% and it is 1.1\% when compared to \emph{drw}. The delay values are extracted using ABC's $print\_stats$ command which calculates and returns single point load independent delay values, hence, these delay values won't be a good representative of the actual delay numbers for a circuit. A better way would be using more advanced delay models and extracting the results post place-and-route.   

Table \ref{Table_run-time} lists run-time values for both training and inference steps. The training times are only applicable to Q-drw and A3C-drw. The infer times for baselines are their normal run-time values, e.g. time it takes to perform drw+map commands. For Q-drw and A3C-drw two run-times, the train times are for the case of 1000 episodes. The training run-times are in the range of minutes to hours and since the training is done only once and it is offline, these values are not too bad. The inference run-time values which are times consumed for actual usage of the framework for synthesizing input circuits are in the order of seconds. Even-though the run-time values of Q-drw and A3C-drw for inference are larger than others, it won't impact the total run-time of synthesis, place, and route, because of the low contribution of synthesis in total run-time of the whole process. Therefore, trading significant improvements in QoR for a few extra seconds on run-time would be totally acceptable. 
%%%%%%%%%%%%%%%%%%%%%%%%%%%%%%%%%%%%%%%%%%%%%%%%%%%%%%%%%%%%%%%%%%%%%%%%%%%%%%%%%%%%%
\begin{table}[t]
\centering
\scriptsize
\caption{Run-time values (\emph{s}) for five different methods: no drw + map: \emph{no rw}, ABC's drw + map: \emph{drw}, 10 consecutive ABC's drw + map: \emph{10 drw}, Q-learning-based rewrite + map: \emph{Q-drw}, and A3C-based rewrite + map: \emph{A3C-drw}. }
\begin{tabular}{@{}cccccccc@{}}
\toprule
	&no drw	 &drw	&	10 drw&	\multicolumn{2}{c}{Q-drw}&		\multicolumn{2}{c}{A3C-drw} \\	
\toprule
circuit	&infer	&infer	&infer		&train	&infer	&train	&infer  \\
\midrule
c1355	&0.17	&0.16	&0.19		&290.24	    &0.58	&690.73	    &1.28  \\
c1908	&0.16	&0.19	&0.19		&235.42 	&1.36	&737.88 	&1.14  \\
c2670	&0.17	&0.17	&0.23		&190.06 	&0.50	&968.66  	&2.04  \\
c3540	&0.19	&0.24	&0.29		&226.34	    &0.62	&1294.72	&2.58  \\
c432	&0.15	&0.15	&0.37		&197.64 	&0.46	&357.11 	&1.09 \\
c499	&0.15	&0.16	&0.19		&187.89	    &0.50	&733.77     &0.99	  \\
c5315	&0.23	&0.27	&0.41		&280.19 	&0.59	&438.40	    &3.51  \\
c6288	&0.36	&0.29	&0.48		&440.62 	&0.73	&539.20   	&4.61  \\
c7552	&0.29	&0.28	&0.45		&486.85 	&0.86	&505.14  	&3.03  \\
c880	&0.15	&0.15	&0.18		&233.81	    &0.46	&432.23     &0.87     \\	
sin	    &0.69	&0.75	&1.34		&674.76 	&1.52	&1258.73	&6.74  \\
max	    &0.29	&0.44	&0.69		&375.42 	&1.08	&666.48 	&3.94  \\
mult	&2.52	&2.45	&5.20		&2708.72	&5.29	&6963.81	&40.95  \\
log2	&3.52	&3.42	&7.21	    &3576.92    &7.33	&35947.66   &30.6  \\
adder	&0.14	&0.15	&0.25		&160.42 	&0.48	&1929.12	&2.16  \\
\bottomrule
average & 0.61	&0.61	&1.17	&684.35	&1.49	&3564.24	&7.03 \\
\label{Table_run-time}
\end{tabular}
\end{table}
%%%%%%%%%%%%%%%%%%%%%%%%%%%%%%%%%%%%%%%%%%%%%%%%%%%%%%%%%%%%%%%%%%%%%%%%%%%%%%%%%%%%%
%%%%%%%%%%%%%%%%%%%%%%%%%%%%%%%%%%%%%%%%%%%%%%%%%%%%%%%%%%%%%%%%%%%%%%%%%%%%%%%%%%%%%
\begin{table}[b]
\centering
\scriptsize
\caption{Results on an start-of-the-art industrial block with 2M+ cells optimized with AISYN using toggle rate as reward. The numbers are normalized to the best results achieved using industry's state-of-the-art synthesis, place-and-route flows. }
\begin{tabular}{@{}cccc@{}}
\toprule
	&	Total net toggle rate &		Combo\_logic power & Total power \\	
\toprule
 AISYN & 0.998 & 0.984 & 0.990\\
\bottomrule
\label{Table_pnr_exp}
\end{tabular}
\end{table}
%%%%%%%%%%%%%%%%%%%%%%%%%%%%%%%%%%%%%%%%%%%%%%%%%%%%%%%%%%%%%%%%%%%%%%%%%%%%%%%%%%%%%

Other than working on open source benchmark circuits, we have used AISYN for experimenting on state-of-the-art industrial level circuit blocks as well. The goal was minimizing power consumption and for this purpose, combinational cones of the target block were extracted and optimized and then stitched back to the original block. In this process, total net toggle rate post synthesis was used as reward in AISYN. Table \ref{Table_pnr_exp} shows the normalized data which is obtained by dividing the numbers to those of a baseline. Baseline results are obtained by using the best of industry's synthesis, place-and-route, and timing tools including Synopsys's Design Compiler and Fusion Compiler, Cadences's Innovus and Tempus.
 The reported Como\_logic and total power numbers in this table are for the postroute stage. In the results that are obtained with AISYN in the loop, we were able to reduce the total power consumption postroute by around 1\% on this important block which is significant given the strength of today's industrial synthesis and place-and-route tools. 
\section{Conclusion}
\label{conc_sec}
In this paper, a superior AI-driven logic synthesis framework is presented. We have used advances in reinforcement learning, deep learning, and representation learning in order to train an agent to learn how to optimize a given logic circuit. This improves on the traditional greedy-based approaches for logic synthesis and offers significant reduction in important quality metrics such as total cell area, total node count, and total edge count. As an example and case study, we added AI awareness to a logic rewriting algorithm to demonstrate the great potentials that AI can bring to the table in logic optimization; our Q-learning-based rewriting algorithm followed by a standard technology mapping could reduce the total cell area by up to 69.3\% compared to a classical rewriting algorithm followed by the same technology mapping command. We also were able to reduce total power consumption postroute by around 1\% on an important industrial benchmark circuit.   
%%%%%%%%%%%%%%%%%%%%%%%%%%%%%%%%%%%%%%%%%%%%%%%%%%%%%%%%%%%%%%%%%%%%%%%%%%%%%%%%%%%%%
\section*{Acknowledgement}
The initial work on AISYN was done from Aug. 2019 till Dec. 2019 during an internship project. The authors would like to thank Ravi Gandikota, Mark Ren, Zeki Bozkus, and Chunhui Li from the NVIDIA corporation for their useful feedback, proof reading of the paper, and great discussions. 

\ifCLASSOPTIONcaptionsoff
  \newpage
\fi
\bibliographystyle{IEEEtran}
\bibliography{IEEEabrv,aisyn}
\end{document}